# Unsupervised Text Extraction from G-Maps

Chandranath Adak
Department of Computer Science and Engineering
University of Kalyani
West Bengal-741235, India
mr.c.adak@ieee.org

*Abstract*—This paper represents an text extraction method from Google maps, GIS maps/images. Due to an unsupervised approach there is no requirement of any prior knowledge or training set about the textual and non-textual parts. Fuzzy C-Means clustering technique is used for image segmentation and Prewitt method is used to detect the edges. Connected component analysis and gridding technique enhance the correctness of the results. The proposed method reaches 98.5% accuracy level on the basis of experimental data sets.

## I. INTRODUCTION

*Text extraction* [1-3] is a technique to take-out the textual portion from a non-textual background, e.g. images of magazines, articles, books, video snapshots, maps, news papers, manuscripts [32], advertisements, banners, street sign boards, name plates, cards, web pages etc. This text extraction problem is challenging in the widespread research areas of computer vision due to its complex backgrounds, i.e. unknown regions: mixture of textual and non-textual graphics.

We deal with text extraction from Google maps (G-Maps), which is very essential for geographical information systems (GIS) [4] to know the information about the particular geographic region, written in the textual portions.

Here we separate the texts from G-Maps in the unsupervised manner, i.e. no prior knowledge is required. We also use *fuzzy c-means (FCM)* clustering [5] for segmenting the input G-Maps, and *Prewitt* [6] method for edge detection. Connected component (CC) analysis and gridding lead the proposed method to be more accurate.

## II. FUZZY C-MEANS CLUSTERING

*Clustering* [7] is a process to partition a space into some regions on the basis of some similarities and properties. FCM clustering technique is developed and improved by Dunn (1973) and Bezdek (1981) respectively. FCM clustering algorithm is as follows:

(i) Randomly choose $K$ points as initial cluster centers $\{C_1, C_2, ..., C_K\}$ from a input space of data set $X \{x_1, x_2, ..., x_n\}$.

(ii) Assign each point to its nearest cluster center, i.e., generate *partition matrix U(X)*.

$$U = [u_{ki}] \; ; \; k=1,2,...K \; ; \; i=1,2,...,n \qquad (1)$$

$u_{ki}$ is the *membership degree* of $i^{th}$ point to $k^{th}$ cluster.

$$u_{ki} = \frac{1}{\sum_{j=1}^{K}\left(\frac{d(z_k, x_i)}{d(z_j, x_i)}\right)^{\frac{2}{(m-1)}}} \; ; \qquad (2)$$

$$1 \leq k \leq K \; ; \; 1 \leq i \leq n$$

$d(z,x)$ is the *distance function* (here we use *Euclidean distance* function); $m(>1)$ is a fuzzy exponent, called as *fuzzifier*.

(iii) $z_k$ is calculated as follows :

$$z_k = \frac{\sum_{i=1}^{n} u_{ki}^m \cdot x_i}{\sum_{i=1}^{n} u_{ki}^m} \qquad (3)$$

(iv) Calculate *cluster validity index* ($J_m$) :

$$J_m = \sum_{k=1}^{K} \sum_{i=1}^{n} u_{ki}^m d^2(z_k, x_i) \qquad (4)$$

Clustering result is better for smaller $J_m$ value.

## III. PROPOSED METHOD

The proposed method consists of following steps:

Step 1: G-Map is taken as input image $I_{RGB}$ (fig.2a).

Step 2: $I_{RGB}$ is converted into grayscale image $I_{gray}$ (fig.2b) and noise is removed.

Step 3: $I_{gray}$ is segmented [6, 8] with the help of FCM, and we get the segmented mask image $I_{mask}$ (fig.2c).

Step 4: $I_e$, i.e. the edge (fig.2d) of $I_{mask}$, is generated using *Prewitt* method [6].

Step 5: Dilation [6, 9] operation is performed on $I_e$ to get the dilated image $I_d$ (fig.2e).

Step 6: We generate all the connected components ($I_{cc}$) [6,10] from $I_d$, by the concept of label matrix. The maximal components $I_{mcc}$ (fig.2f) are extracted based on a threshold [6, 11] ($T : 0<T<mn; \; [m \; n] = size(I_{gray})$).

Step 7: Now we have to remove the attached non-textual parts from $I_{mcc}$. In most of the times, the texts are written with some curvy, thin, smaller lines. We are gridding [7] (fig.2g) the $I_{mcc}$, and partitioning into *3X3* block (fig.1). The value of $p_i$ *(i=1,2,...,9)* is either *0* or *1* due to binarization. If a total row, column, or diagonal is *1*, then it has a high possibility that the block is under the non-textual part. After *3X3* gridding, the image may be revised with *5X5* gridding. For $p_i=0$, this is a background pixel.

| $p_1(x-1, y-1)$ | $p_2(x-1, y)$ | $p_3(x-1, y+1)$ |
| --- | --- | --- |
| $p_4(x, y-1)$ | **$p_5(x, y)$** | $p_6(x, y+1)$ |
| $p_7(x+1, y-1)$ | $p_8(x+1, y)$ | $p_9(x+1, y+1)$ |

*Fig.1.* 3X3 Grid Structure Element (GSE)

*Steps 6,7* are repeated (with changing the threshold value : *T* ) for a satisfactory outcome ($I_O$).

Step 8: We regenerate the gray-scale image ($I_f$) from $I_O$ as follows :

// $BG_{color}$ = background color
// (x,y) is the image pixel co-ordinate value

if ($I_O(x,y)= = 1$)
 $I_f(x,y)=I_{gray}(x,y)$ ;
else
 $I_f(x,y)=BG_{color}$ ;

$I_f$ is the text portion (fig.2h), separated from the G-Map. It is the final output of our proposed method.

## IV. EXPERIMENTAL RESULTS AND COMPARISONS

To assess the immovability and accurateness of the proposed technique, the results are obtained from different G-Maps (fig. 2-3) and compared with other existing methods.

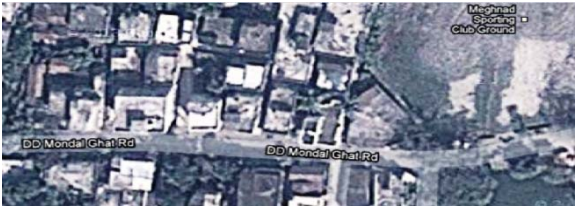

*Fig. 2(a)*

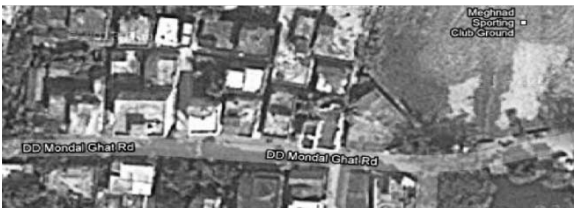

*Fig. 2(b)*

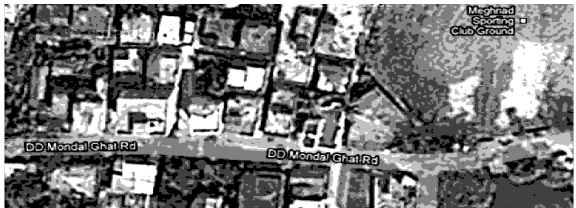

*Fig. 2(c)*

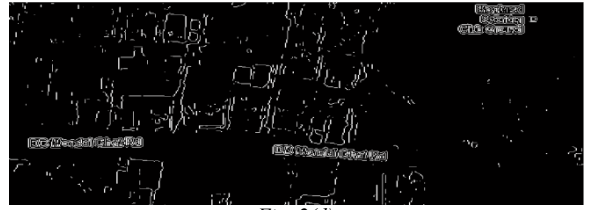

*Fig. 2(d)*

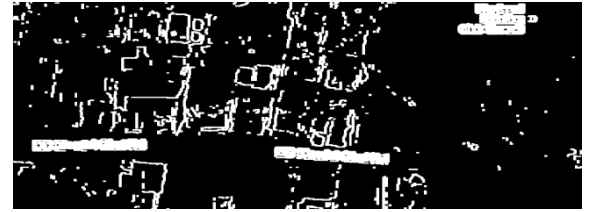

*Fig. 2(e)*

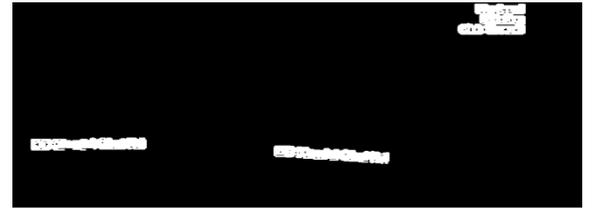

*Fig. 2(f)*

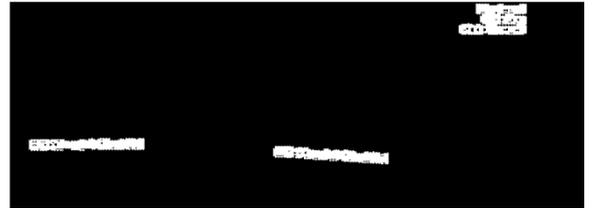

*Fig. 2(g)*

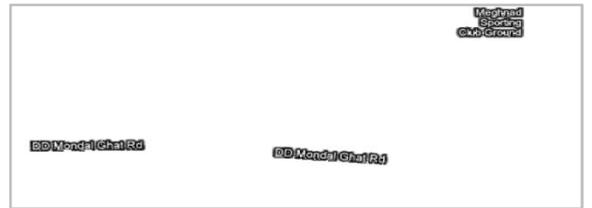

*Fig. 2(h)*

*Fig.2.* Google Map Satellite image (400X600) of *DD Mondal Ghat Road, Kolkata* (22.660411°N, 88.360838°E) : (a) original, (b) gray, (c) segmented, (d) edge, (e) dilated, (f) maximal CC (T=2000), (g) after gridding, (h) extracted text.

From the fig.2, it is cleared that we are able to extract the textual parts from the G-Maps and the proposed method gives noise-free fine results. But, in fig.2, we are unable to see the utility of grid structuring technique (as step-7). It is shown in fig.3.

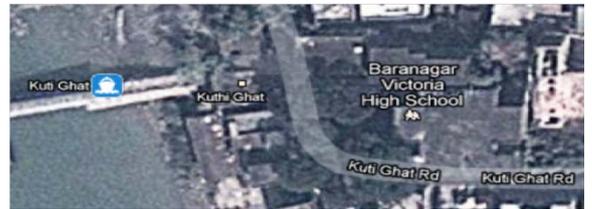

*Fig. 3(a)*

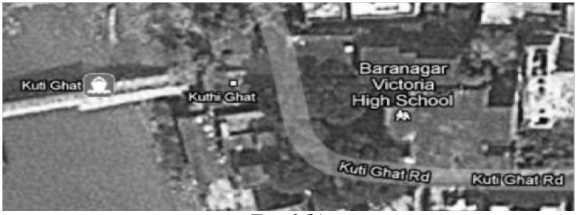
*Fig. 3(b)*

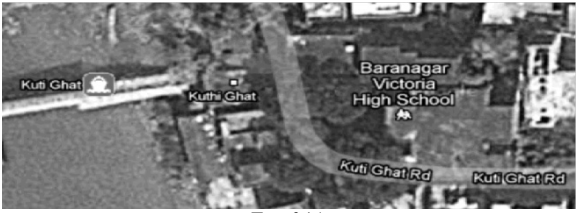
*Fig. 3(c)*

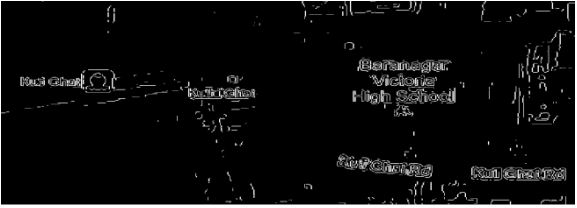
*Fig. 3(d)*

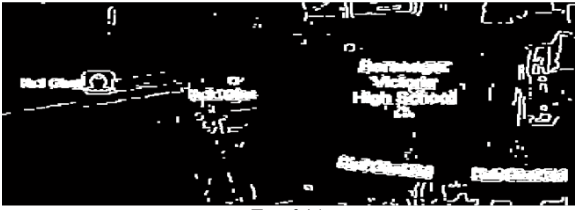
*Fig. 3(e)*

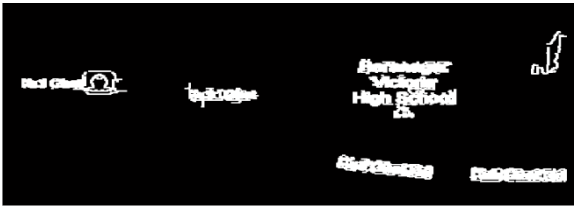
*Fig. 3(f)*

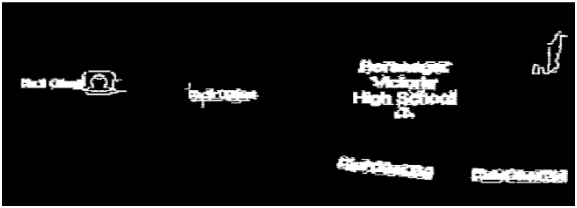
*Fig. 3(g)*

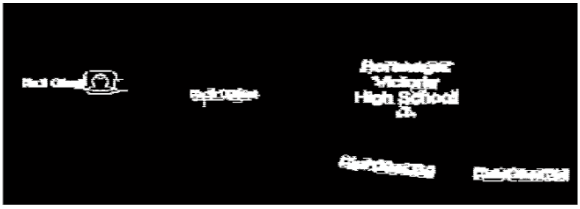
*Fig. 3(h)*

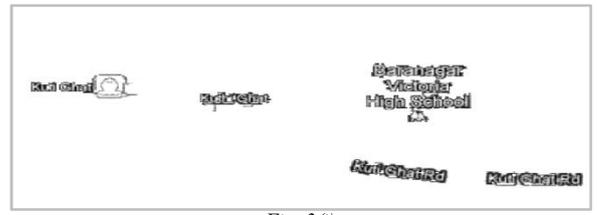
*Fig. 3(i)*

*Fig.3.* Google Map Satellite image (400X600) of *Kuti Ghat, Kolkata* (22.635184 ºN, 88.362865 ºE) : (a) original, (b) gray, (c) segmented, (d) edge, (e) dilated, (f) maximal CC (T=400), (g) gridding, (h) maximal CC, after gridding (T=410), (i) extracted text.

In fig.3, we can see the effectiveness of grid structuring technique, where the non-textual part is easily removed to get the proper output.

To measure the accuracy of the proposed method, the *confusion matrix* [33] is drawn as fig.4.

*Fig.4.* Confusion Matrix

Accuracy = (97+100)/(97+100+0+3) = 197/200 = 0.985, Accuracy (%) = **98.5%**

TABLE I.  COMPARISON OF RESULTS

| Sl. No. | Author(s) / Method | Accuracy (%) |
|---|---|---|
| 1. | S. Grover et al. [12] | 99 |
| 2. | P. Nagabhushan et al. [13] | 97.12 |
| 3. | S. A. Angadi et al. [14] | 96.60 |
| 4. | W. Boussellaa et al. [15] | 96 |
| 5. | S. Zhixin et al. [16] | 95 |
| 6. | S. Audithan et al. [17] | 94.80 |
| 7. | Keechul Jung et al. [18] | 92.38 |
| 8. | Davod Zaravi et al. [19] | 91.20 |
| 9. | Shivakumara P. et al. [20] | 89.67 |
| 10. | J. Fabrizio et al. [21] | 88.83 |
| 11. | Miriam Leon et al. [22] | 86.35 |
| 12. | W. A. L. Alves et al. [23] | 85.93 |
| 13. | M. Leon et al. [24] | 85.78 |
| 14. | Sunil Kumar et al. [25] | 85.54 |
| 15. | T. Q. Phan et al. [26] | 84.90 |
| 16. | V. Vijayakumar et al. [27] | 84.89 |
| 17. | Y. Zhan et al. [28] | 84.30 |
| 18. | G.R.Mohan Babu et al. [29] | 84.01 |
| 19. | Y. F. Pan et al. [30] | 83.44 |
| 20. | Proposed Method | **98.5** |

From the *table-I*, it is easily observed that the proposed method has better result than some other methods [31].

V. CONCLUSION AND FUTURE WORK

The proposed hybrid model is tested on two hundreds test samples and we achieve 98.5% accuracy. Due to connected component analysis, there are some possibilities to get the non-textual portion touched with the textual ones. This problem is overcome by the grid structuring scrutiny. The proposed method also works well in topological maps, historical maps, GIS maps etc., and it is language independent of the maps. The limitation of this proposed method is that it is not fully automatic due to thresholding and selection of better result is dependent on human eye. So our next venture will be overcome this limitation to make this system more accurate.

ACKNOWLEDGMENT

I would like to heartily thank *Prof. Bidyut B. Chaudhuri*, IEEE Fellow, Head, Computer Vision and Pattern Recognition Unit, Indian Statistical Institute, Kolkata 700108, India, for discussion various aspects of this research field.